\documentclass[conference]{IEEEtran}
\IEEEoverridecommandlockouts
\usepackage[utf8]{inputenc}
\usepackage[T1]{fontenc}
\usepackage{cite}
\usepackage{amsmath,amssymb,amsfonts}
\usepackage{algorithmic}
\usepackage{graphicx}
\usepackage{adjustbox}
\usepackage{textcomp}
\usepackage{xcolor}
\usepackage{array}
\usepackage{booktabs}
\def\BibTeX{{\rm B\kern-.05em{\sc i\kern-.025em b}\kern-.08em
    T\kern-.1667em\lower.7ex\hbox{E}\kern-.125emX}}
\begin{document}

\title{C-GRASP: Clinically-Grounded Reasoning for Affective Signal Processing}

\author{\IEEEauthorblockN{1\textsuperscript{st} Cheng Lin Cheng}
\IEEEauthorblockA{\textit{Department of Physics} \\
\textit{National Central University}\\
Taoyuan, Taiwan \\
}
\and
\IEEEauthorblockN{2\textsuperscript{nd} Ting Chuan Lin}
\IEEEauthorblockA{\textit{Department of Physics} \\
\textit{National Central University}\\
Taoyuan, Taiwan \\
}
\and
\IEEEauthorblockN{3\textsuperscript{rd} Chai Kai Chang}
\IEEEauthorblockA{\textit{Center for Education} \\
\textit{National Central University}\\
Taoyuan, Taiwan \\
ckchang@ncu.edu.tw\\
\thanks{Corresponding author: ckchang@ncu.edu.tw}}
}

\maketitle

\begin{abstract}
Heart rate variability (HRV) is a pivotal non-invasive marker for autonomic monitoring; however, applying Large Language Models (LLMs) to HRV interpretation is hindered by physiological hallucinations, where models struggle with respiratory sinus arrhythmia (RSA) contamination, short-data instability in nonlinear metrics, and the neglect of individualized baselines in favor of population norms. We propose C-GRASP (Clinically-Grounded Reasoning for Affective Signal Processing), a guardrailed RAG-enhanced pipeline that decomposes HRV interpretation into eight traceable reasoning steps. Central to C-GRASP is a Z-score Priority Hierarchy that enforces the weighting of individualized baseline shifts over normative statistics. The system effectively mitigates spectral hallucinations through automated RSA-aware guardrails, preventing contamination of frequency-domain indices. Evaluated on 414 trials from the DREAMER dataset, C-GRASP integrated with high-scale reasoning models (e.g., MedGemma3-thinking) achieved superior performance in 4-class emotion classification (37.3\% accuracy) and achieved a Clinical Reasoning Consistency (CRC) score of 69.6\%. Ablation studies confirm that the individualized Delta Z-score module serves as the critical logical anchor, preventing the ``population bias'' common in native LLMs. Ultimately, C-GRASP transitions affective computing from black-box classification to transparent, evidence-based clinical decision support, paving the way for safer AI integration in biomedical engineering.
\end{abstract}


\section{Introduction}

Heart rate variability (HRV) is a key non-invasive marker for autonomic monitoring, yet applying Large Language Models (LLMs) to HRV interpretation poses unique challenges: respiratory sinus arrhythmia (RSA) can contaminate frequency-domain indices, short data segments destabilize nonlinear metrics, and visual plots are prone to scaling-induced hallucinations. A critical gap exists in current LLM-based HRV interpretation: models lack respect for individualized baselines, defaulting to population norms. This limitation leads to misdiagnosis in high-HRV populations, where individuals with naturally elevated HRV metrics are incorrectly flagged as abnormal when their values exceed population averages, despite being normal relative to their personal baselines.

We propose \textbf{C-GRASP} (\textbf{C}linically-\textbf{G}rounded \textbf{R}easoning for \textbf{A}ffective \textbf{S}ignal \textbf{P}rocessing), a guardrailed RAG-enhanced pipeline that decomposes HRV interpretation into eight traceable steps. The key innovation of C-GRASP is the integration of a \textbf{Dual Z-score Priority Hierarchy} with \textbf{quantitative guardrails} into the RAG reasoning chain architecture, ensuring that individualized baseline shifts take precedence over normative statistics while automatically detecting and mitigating spectral artifacts. Key contributions:
\begin{itemize}
    \item \textbf{Stepwise reasoning} (Steps 1--8) for transparent, auditable interpretation.
    \item \textbf{Quality-aware guardrails} triggered by quantitative features (RSA severity, data length) to prevent HRV pitfalls.
    \item \textbf{Evidence governance} in RAG: re-ranking by metric reliability and study design.
    \item \textbf{Dual Z-score normalization} for directional consistency in within-subject tracking.
\end{itemize}

\section{Related Work}

\textbf{Methodological Challenges in HRV Analysis.} 
Traditional interpretation of physiological signals faces significant pitfalls due to motion and physiological artifacts. As demonstrated in our previous work on EEG sleep staging~\cite{cheng2025}, artifact removal via MNE-based ICA can significantly impact classification performance, yielding F1 score gains of up to +7.3\% in sensitive stages like REM while potentially suppressing intrinsic rhythmic patterns in deeper stages like N3. Such variability necessitates the rigorous, quality-aware reasoning implemented in C-GRASP. Frequency-domain metrics like LF/HF are often confounded by respiratory sinus arrhythmia (RSA) and are unreliable proxies for sympatho-vagal balance~\cite{billman2013,heathers2014}. Similarly, nonlinear indices such as Sample Entropy (SampEn) are highly sensitive to data length and parameter selection, leading to inconsistent results in short-term recordings~\cite{zhao2015}. These methodological limitations necessitate rigorous quality control, which is often absent in automated systems.

\textbf{LLMs in Physiological Reasoning.}
Large Language Models (LLMs) have demonstrated potential in clinical report generation; however, their application to physiological data is hindered by ``normative bias''---a tendency to prioritize population statistics over individualized baselines---and a susceptibility to numerical hallucinations. Unlike general medical QA tasks, physiological interpretation requires precise handling of quantitative contradictions (e.g., high variability vs. low complexity), a capability often lacking in standard retrieve-and-generate frameworks.

\textbf{Guardrailed RAG Systems.}
Retrieval-Augmented Generation (RAG) grounds LLM outputs in external knowledge~\cite{pikerag}, yet generic RAG implementations often fail to enforce domain-specific constraints. C-GRASP addresses these gaps by integrating a \textbf{quality-aware guardrail system} that dynamically modulates retrieval based on signal quality (e.g., RSA severity) and enforces a strict Z-score priority hierarchy, ensuring that clinical reasoning is both evidence-based and physiologically valid.

\section{Methods}

\subsection{System Overview}

C-GRASP is a guardrailed RAG-enhanced stepwise reasoning pipeline for clinically traceable HRV interpretation. The system comprises four core modules:
\begin{enumerate}
    \item Feature Construction with individualized normalization,
    \item RAG-based clinical knowledge retrieval with evidence governance,
    \item Stepwise reasoning (Steps 1--7) with quality-aware guardrails, and
    \item Template-constrained integration (Step 8) with post-processing validation.
\end{enumerate}

\begin{figure*}[htbp]
\centering
\includegraphics[width=0.9\linewidth]{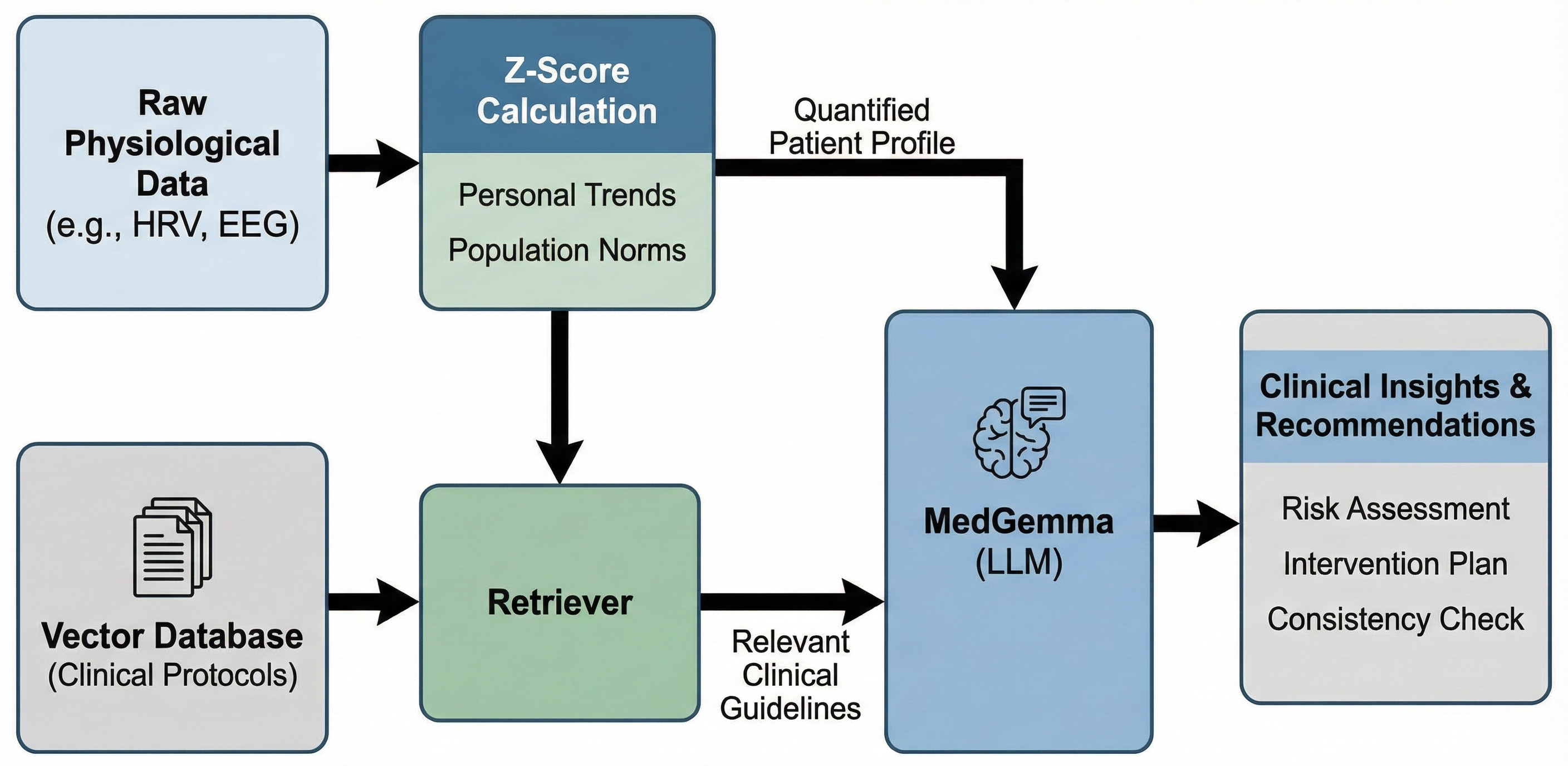}
\caption{The C-GRASP System Architecture. The framework integrates individualized feature normalization, dynamic RAG retrieval, and guardrailed stepwise reasoning to generate clinically traceable reports.}
\label{fig:system_arc}
\end{figure*}

\subsection{Input Features and Individualized Normalization}

\subsubsection{Core HRV Features}
The system receives preprocessed HRV features following standard guidelines\footnote{Preprocessing details: 4 Hz cubic spline interpolation; Smoothness priors detrending ($\lambda=500$); Frequency bands: ULF ($<0.04$ Hz), LF ($0.04$--$0.15$ Hz), HF ($0.15$--$0.40$ Hz); SampEn ($m=2, r=0.2\sigma$); DFA scales ($4$--$16$ beats).}. Table~\ref{tab:hrv-features} categorizes the features by domain. Each feature is accompanied by domain-specific quality indicators computed during preprocessing.

\begin{table}[htbp]
\caption{HRV Feature Categories and Metrics}
\begin{center}
\resizebox{\columnwidth}{!}{
\begin{tabular}{|>{\centering\arraybackslash}m{2.5cm}|m{6.0cm}|}
\hline
\textbf{Domain} & \textbf{Metrics} \\
\hline
Time-domain & MeanRR, SDNN (total variability), MeanHR, SDHR, RMSSD (parasympathetic marker), NN50/pNN50, SDNN index \\
\hline
Frequency-domain & Peak frequencies: ULF, LF, HF; Power ratios: ULF\_ratio, LF\_ratio, HF\_ratio, LF/HF \\
\hline
Nonlinear/Geometric & Poincaré axes: SD1, SD2; Sample Entropy (SampEn); DFA scaling exponent (DFA$_\alpha$) \\
\hline
\end{tabular}
}
\label{tab:hrv-features}
\end{center}
\end{table}

\subsubsection{Dual Z-Score Normalization}
To address the well-known issue of traditional Z-score contradicting delta direction (e.g., $z_{\text{trad}} > 0$ but $\Delta < 0$), we introduce a \textbf{dual Z-score mechanism}. The traditional Z-score and Delta Z-score are defined as
\begin{equation}
    z_{\text{trad}} = \frac{x_{\text{stim}} - \mu_{\text{pop}}}{\sigma_{\text{pop}}},
    \label{eq:trad-z}
\end{equation}
and
\begin{equation}
    z_{\Delta} = \frac{\Delta x - \mu_{\Delta}}{\sigma_{\Delta}}, \quad \Delta x = x_{\text{stim}} - x_{\text{baseline}},
    \label{eq:delta-z}
\end{equation}
respectively. The Delta Z-score ensures directional consistency: $\Delta > 0$ implies $z_{\Delta} > 0$, eliminating sign contradictions in within-subject comparisons. The system prioritizes Delta Z-scores for within-subject analysis while retaining traditional Z-scores for population-level context.

\textbf{Formal role of $z_{\Delta}$ in Step 6.} To reduce within-subject logical conflicts in Step 6 (within-subject profiling), we explicitly define the primary within-subject evidence as $z_{\Delta}$:
\begin{equation}
    z_{\text{S6}}(x) = z_{\Delta}(x),
    \label{eq:z-s6}
\end{equation}
and only treat $z_{\text{trad}}$ (\eqref{eq:trad-z}) as auxiliary population context. When the two sources disagree in direction, we flag a potential reasoning conflict and force Step 6 to follow $z_{\Delta}$:
\begin{equation}
    \mathbb{1}_{\text{conflict}} = \mathbb{1}\!\left[\mathrm{sign}\!\left(z_{\text{trad}}\right) \neq \mathrm{sign}\!\left(z_{\Delta}\right)\right].
    \label{eq:z-conflict}
\end{equation}

\subsubsection{Change Analysis Metrics}
Within-subject changes are quantified as absolute deltas ($\Delta_{\text{feat}} = x_{\text{stim}} - x_{\text{baseline}}$) and percentage changes ($\Delta_{\text{feat\_pct}}$). Core delta metrics include RMSSD, SDNN, MeanHR, SampEn, and DFA$_\alpha$, enabling longitudinal tracking of autonomic shifts relative to individual baselines.

\subsubsection{Multi-Modal Features}
When available, we integrate supplementary modalities (e.g., EEG spectral bands and ECG-derived respiration). EEG features are normalized with device-specific scaling awareness to prevent cross-device artifacts.

\subsection{RAG-Based Clinical Knowledge Retrieval}

We implement retrieval using the PIKE-RAG framework~\cite{pikerag}, which emphasizes domain knowledge extraction, knowledge organization, and task decomposition to support multi-step, traceable reasoning beyond retrieve-and-stuff baselines.

\subsubsection{Vector Store Architecture}
Clinical PDF documents are parsed, chunked using recursive character splitting (preserving sentence boundaries), and embedded via a biomedical sentence encoder (BioLORD-2023). The resulting embeddings are stored in a \textbf{Chroma} vector database with persistent storage, enabling efficient similarity search via approximate nearest neighbors. Each chunk retains source metadata (file name, page number, study design label) for downstream evidence attribution. Table~\ref{tab:rag-params} summarizes the key RAG configuration parameters.

\begin{table}[htbp]
\caption{RAG System Configuration Parameters}
\begin{center}
\resizebox{\columnwidth}{!}{
\begin{tabular}{|>{\centering\arraybackslash}m{2.8cm}|>{\centering\arraybackslash}m{2.2cm}|>{\centering\arraybackslash}m{3.2cm}|}
\hline
\textbf{Parameter} & \textbf{Value} & \textbf{Description} \\
\hline
\multicolumn{3}{|c|}{\textit{Document Processing}} \\
\hline
Chunk size & 1000 tokens & Text segment length \\
\hline
Chunk overlap & 200 tokens & Overlap for context continuity \\
\hline
Embedding model & BioLORD-2023 & Biomedical sentence encoder \\
\hline
\multicolumn{3}{|c|}{\textit{Retrieval Settings}} \\
\hline
Top-$k$ passages & $k=5$ & Retrieved candidates per query \\
\hline
Similarity threshold & $\tau=0.3$ & Minimum relevance score \\
\hline
\multicolumn{3}{|c|}{\textit{Metric Reliability Weights ($\beta_{\text{metric}}$)}} \\
\hline
RMSSD & 0.9 & High: reliable parasympathetic \\
\hline
SDNN & 0.7 & Medium: requires HR correction \\
\hline
SampEn & 0.6 & Medium: parameter-sensitive \\
\hline
DFA$_\alpha$ & 0.5 & Low-medium: not direct ANS \\
\hline
SD1/SD2 & 0.4 & Low: geometric, not complexity \\
\hline
LF/HF & 0.3 & Low: unreliable balance proxy \\
\hline
\multicolumn{3}{|c|}{\textit{Study Design Modifiers ($\gamma_{\text{design}}$)}} \\
\hline
RCT / Controlled trial & 1.08 & Boost high-quality evidence \\
\hline
Clinical observational & 1.05 & Slight boost \\
\hline
Opinion / Commentary & 0.97 & Slight penalty \\
\hline
Threshold-heavy passage & 0.85 & 15\% penalty for abs.\ thresholds \\
\hline
\end{tabular}
}
\label{tab:rag-params}
\end{center}
\end{table}

Rather than using raw similarity scores, we implement an \textbf{Evidence Governance} mechanism with domain-specific weighting:
\begin{equation}
    s_{\text{adj}} = s_{\text{raw}} \times w_{\text{domain}}
\end{equation}
where $w_{\text{domain}}$ aggregates metric reliability, study-design modifiers, and a penalty for passages over-relying on population thresholds (see Table~\ref{tab:rag-params}). In particular, the metric reliability weights in Table~\ref{tab:rag-params} directly instantiate $\beta_{\text{metric}}$ in \eqref{eq:w-domain}.

A strict \textbf{Decision Hierarchy} is enforced: within-subject Z-scores take precedence over complexity metrics, which in turn take precedence over absolute values and, lastly, over literature norms. Retrieved passages are ranked by adjusted similarity scores and provided to the LLM with explicit source attribution.

\subsubsection{Clinical Knowledge Base}
The RAG corpus comprises curated clinical/methodological literature, each assigned a primary ``responsibility'' for safe HRV reasoning:

\begin{itemize}
    \item \textbf{Frequency-domain limitations and LF/HF critique}~\cite{billman2013,heathers2014}: Core critique that LF/HF is not a reliable sympatho-vagal balance proxy; highlights dynamic-context distortions and short-term frequency-domain pitfalls (windowing, respiration, protocol dependence).
    
    \item \textbf{Heart-rate correction and repeatability}~\cite{gasior2016}: Emphasizes the impact of mean HR and respiration on HRV repeatability; supports HR-corrected interpretation for RMSSD/SDNN.
    
    \item \textbf{Normative values}~\cite{gasior2018} (context only; never overriding within-subject baselines): Pediatric reference ranges with mean-HR correction; used strictly for contextual background when individualized baselines are unavailable.
    
    \item \textbf{Mixed autonomic states / co-activation}~\cite{eickholt2018,berg2011}: Clinical and experimental evidence for simultaneous sympathetic/parasympathetic activation leading to non-linear cardiovascular responses; motivates ``dual activation'' warnings.
    
    \item \textbf{Nonlinear/fractal analysis (DFA) interpretability}~\cite{echeverria2003}: DFA local-scaling exponent profiling; supports guardrails around scale-range ambiguity in short HRV segments.
    
    \item \textbf{Entropy parameterization and data-length sensitivity}~\cite{mayer2014,zhao2015}: Practical guidance for SampEn/FuzzyEn parameter choices ($m$, $r$) and minimum data-length considerations; clinical case-control examples emphasizing parameter/data-length coupling.
    
    \item \textbf{Geometric indices under intervention}~\cite{vanderlei2019}: Exercise intervention effects on SD1/SD2 and geometric indices; used to contextualize geometric changes while avoiding over-claiming ``complexity'' from geometry alone.
\end{itemize}

\subsubsection{Data-Driven Dynamic Query Adaptation}
Rather than using static retrieval queries, our system dynamically adapts RAG strategies based on the input HRV data through three mechanisms:

\textbf{(1) Z-Score-Aware Query Construction.} For each HRV metric, we first classify relative changes using:
\begin{equation}
    \text{state}(x) = 
    \begin{cases}
        \text{marked} & |z_{\Delta}| \geq 2.0 \\
        \text{moderate} & 1.0 \leq |z_{\Delta}| < 2.0 \\
        \text{mild} & 0.5 \leq |z_{\Delta}| < 1.0 \\
        \text{baseline} & |z_{\Delta}| < 0.5
    \end{cases}
\end{equation}
where $z_{\Delta}$ is the within-subject Delta Z-score (\eqref{eq:delta-z}). If Z-scores are unavailable, we fall back to absolute thresholds with reduced confidence. The detected states are then embedded into retrieval queries (e.g., ``RMSSD elevated vagal tone'' vs.\ ``RMSSD reduced parasympathetic withdrawal'').

\textbf{(2) Contradiction-Triggered Warning Queries.} When physiologically inconsistent patterns are detected, the system automatically injects targeted warning queries to retrieve relevant methodological literature. Table~\ref{tab:contradiction} summarizes the contradiction detection rules.

\begin{table}[htbp]
\caption{Contradiction Detection Rules for Dynamic Query Injection}
\begin{center}
\resizebox{\columnwidth}{!}{
\begin{tabular}{|>{\centering\arraybackslash}m{3.5cm}|>{\centering\arraybackslash}m{4.5cm}|}
\hline
\textbf{Detected Pattern} & \textbf{Injected Query Topic} \\
\hline
RMSSD$\uparrow$ + LF/HF$\uparrow$ & Sympathetic-parasympathetic coactivation \\
\hline
HR$\uparrow$ + SampEn$\downarrow$ + LF/HF$\downarrow$ & LF/HF unreliability, respiratory confound \\
\hline
DFA$_\alpha$$\uparrow$ + RMSSD$\downarrow$ & DFA not a direct autonomic measure \\
\hline
SD1/SD2$\downarrow$ + SampEn$\downarrow$ & Geometric vs.\ complexity construct difference \\
\hline
LF/HF $> 3.0$ or $< 0.3$ & Extreme ratio, respiratory artifact \\
\hline
\end{tabular}
}
\label{tab:contradiction}
\end{center}
\end{table}

\textbf{(3) Multi-Factor Domain Weighting.} Retrieved passages are re-ranked using a composite weight:
\begin{equation}
    w_{\text{domain}} = w_{\text{base}} \times (1 + \alpha_{\text{topic}}) \times (1 + \beta_{\text{metric}}) \times \gamma_{\text{design}}
    \label{eq:w-domain}
\end{equation}
where $\alpha_{\text{topic}}$ rewards topic overlap with query, $\beta_{\text{metric}}$ scales by metric reliability weights listed in Table~\ref{tab:rag-params} (i.e., for each retrieved passage we set $\beta_{\text{metric}}$ to the value associated with its primary HRV metric; e.g., RMSSD $\rightarrow 0.9$, SDNN $\rightarrow 0.7$, LF/HF $\rightarrow 0.3$), and $\gamma_{\text{design}}$ adjusts for study design (RCT: 1.08, opinion: 0.97). Passages over-relying on absolute thresholds (e.g., ``RMSSD $>$ 40 ms indicates...'') receive an additional 15\% penalty.

\subsection{Quantitative Image Features: Preventing Visual Hallucinations}

A critical innovation of our system is the inclusion of \textbf{quantitative image-derived features} extracted directly from the underlying data series, independent of image rendering. This design prevents LLM hallucinations caused by image scaling, compression artifacts, or visual interpretation errors.

\subsubsection{Poincaré Plot Quantitative Features}
For the Poincaré plot (RRI(n) vs. RRI(n+1)), we extract quantitative statistics from point coordinates (e.g., boundary stats, density center, and point count). The \textbf{scatter point count} ($N_{\text{poincare}}$) is a guardrail trigger: when $N_{\text{poincare}} < 100$, nonlinear interpretation is disabled due to estimation instability.

\subsubsection{Power Spectral Density (PSD) Quantitative Features}
For frequency-domain analysis, we extract quantitative features from raw PSD (band powers and peak frequencies). For respiratory coupling, we compare the quantitative respiratory frequency ($f_{\text{resp}}$) against the HF peak frequency ($f_{\text{HF}}$) and trigger RSA contamination when $|f_{\text{resp}} - f_{\text{HF}}| < 0.05$ Hz.

\subsubsection{Visual Inputs and Multi-Modal Reasoning}
While quantitative features provide numerical grounding, we also use visualization panels (Poincaré, signal quality, PSD) for cross-checking. If visual impressions contradict quantitative features, we flag potential hallucinations.

\subsection{Signal Quality and Guardrail Gating}
C-GRASP employs multi-dimensional quality indicators (e.g., artifact rate, valid RR ratio, spectral reliability, respiratory contamination, and nonlinear stability) to trigger system-level guardrails. The rationale for this gating mechanism is grounded in our prior findings~\cite{cheng2025}, which quantified that while targeted denoising enhances specific segments (e.g., +5.4\% F1 gain for N1), it can also introduce signal distortions in others. Therefore, Step 1 (Signal Quality) explicitly gates downstream analysis when artifact rates exceed 0.2 to ensure interpretation reliability.

\subsection{Within-Subject Profiling}

For each participant, baseline statistics are computed on-the-fly during dataset initialization as a per-subject reference distribution: $\mu_{\text{baseline}} = \text{mean}(\{x_{\text{trial}_i}\})$ and $\sigma_{\text{baseline}} = \text{std}(\{x_{\text{trial}_i}\})$ for each feature. This enables Step 6 to distinguish transient emotional shifts from chronic baseline shifts by comparing the current trial against the participant's own longitudinal profile, rather than relying solely on population norms.

\textbf{Data leakage consideration (retrospective baseline).} Our long-term deployment target is wearable-based monitoring, where a subject-specific baseline is accumulated from that individual's historical measurements. In this offline study, we use all available trials from the same subject to \textit{simulate} a long-term tracking scenario and build a stable individualized norm. This baseline construction is \textit{unsupervised} (it does not use labels), but it is non-causal in the strict sense because the current trial (and potentially future trials) may contribute to the estimated baseline distribution. Therefore, results that depend on within-subject baselines should be interpreted as a \textit{retrospective} setting and may be slightly optimistic. In future work and real-world deployment, the baseline will be computed causally using only prior windows/sessions (e.g., chronological accumulation or leave-one-trial-out baselines) to eliminate this concern.

\subsection{Stepwise Reasoning Pipeline (Steps 1--7)}

The reasoning process is decomposed into seven specialized steps, each with defined inputs, tasks, and outputs, as summarized in Table~\ref{tab:steps}. The decomposition follows a \textbf{domain-driven design}: each step addresses a distinct physiological domain, enabling targeted guardrails and interpretable intermediate outputs.

\textbf{Design Rationale.}
\begin{itemize}
    \item \textbf{Step 1 (Signal Quality)} gates downstream analysis: if artifact rate $>0.2$ or valid RR ratio $<0.8$, subsequent steps receive explicit quality warnings.
    \item \textbf{Step 2 (Time-Domain)} provides the most reliable autonomic markers (RMSSD, SDNN) with lowest methodological controversy.
    \item \textbf{Step 3 (Frequency-Domain)} is where RSA contamination is most dangerous; guardrails are mandatory here.
    \item \textbf{Step 4 (Nonlinear)} requires sufficient data length; data-length guardrails prevent spurious entropy/DFA claims.
    \item \textbf{Steps 5--6 (Delta \& Within-Subject)} enforce individualized interpretation, prioritizing $z_{\Delta}$ over population norms.
    \item \textbf{Step 7 (EEG)} is optional and only activated when multi-modal data is available.
\end{itemize}

\begin{table}[htbp]
\caption{Stepwise Reasoning Pipeline Overview}
\begin{center}
\resizebox{\columnwidth}{!}{
\begin{tabular}{|>{\centering\arraybackslash}m{0.6cm}|>{\centering\arraybackslash}m{1.8cm}|>{\centering\arraybackslash}m{4.0cm}|>{\centering\arraybackslash}m{2.3cm}|}
\hline
\textbf{Step} & \textbf{Name} & \textbf{Key Tasks} & \textbf{Output} \\
\hline
1 & Signal Quality & Evaluates artifact rate, valid RR ratio, quality scores (0--1) & Quality grade, recommendations \\
\hline
2 & Time-Domain & Analyzes RMSSD, SDNN, MeanHR with Z-scores & Vagal tone, arousal assessment \\
\hline
3 & Frequency-Domain + RSA Guardrails & Analyzes PSD, LF/HF using quantitative features; RSA guardrail: $|f_{\text{resp}} - f_{\text{HF}}| < 0.05$ Hz & Frequency analysis, RSA warnings \\
\hline
4 & Nonlinear + Data Length Guardrails & Analyzes SampEn, DFA$_\alpha$, Poincaré (SD1, SD2); Guardrail: $N_{\text{poincare}} < 100$ & Nonlinear metrics, stability warnings \\
\hline
5 & Baseline Delta & Computes within-subject changes; prioritizes Delta Z-scores & Change magnitude, direction \\
\hline
6 & Within-Subject Profile & Compares trial against subject's baseline statistics & Longitudinal context \\
\hline
7 & EEG Integration & Integrates EEG spectral features (alpha/beta) when available & Multi-modal assessment \\
\hline
\end{tabular}
}
\label{tab:steps}
\end{center}
\end{table}

\subsection{Guardrails: Quality-Aware Gating}

Guardrails operate at the system level to prevent known HRV interpretation pitfalls. Table~\ref{tab:guardrails} summarizes the trigger conditions and actions.

\begin{table}[htbp]
\caption{Guardrail Trigger Conditions and Actions}
\begin{center}
\resizebox{\columnwidth}{!}{
\begin{tabular}{|>{\centering\arraybackslash}m{1.6cm}|>{\centering\arraybackslash}m{2.8cm}|>{\centering\arraybackslash}m{3.2cm}|}
\hline
\textbf{Guardrail} & \textbf{Trigger Condition} & \textbf{System Action} \\
\hline
RSA severe & $|f_{\text{resp}} - f_{\text{HF}}| < 0.05$ Hz & Prohibit LF/HF; rewrite prompt \\
\hline
RSA moderate & $0.05 \leq |\cdot| < 0.08$ Hz & Caution; guarded interpretation \\
\hline
{\shortstack{Nonlinear\\reliability}} & $N_{\text{poincare}} < 100$ & Prohibit SampEn/DFA use \\
\hline
ULF dominance & ULF ratio $> 0.5$ & Warn frequency unreliability \\
\hline
\end{tabular}
}
\label{tab:guardrails}
\end{center}
\end{table}

\subsubsection{RSA Severity Grading and Dynamic Prompt Adjustment}
Rather than a binary RSA flag, we implement a \textbf{four-level severity grading} based on quantitative frequency alignment:
\begin{equation}
    \text{severity} = 
    \begin{cases}
        \text{severe} & |f_{\text{resp}} - f_{\text{HF}}| < 0.05~\text{Hz} \\
        \text{moderate} & 0.05 \leq |f_{\text{resp}} - f_{\text{HF}}| < 0.08~\text{Hz} \\
        \text{mild} & 0.08 \leq |f_{\text{resp}} - f_{\text{HF}}| < 0.12~\text{Hz} \\
        \text{none} & \text{otherwise}
    \end{cases}
    \label{eq:rsa-severity}
\end{equation}
When \textbf{severity = severe}, the Step 3 system prompt is dynamically rewritten to:
\begin{enumerate}
    \item Explicitly prohibit interpreting high HF power as strong parasympathetic activity.
    \item Prohibit using LF/HF ratio for sympathovagal balance assessment.
    \item Redirect reasoning to time-domain metrics (RMSSD, SDNN) as primary evidence.
\end{enumerate}
For \textbf{severity = moderate}, the prompt issues a caution rather than prohibition, allowing guarded interpretation with explicit acknowledgment of RSA influence.

\subsubsection{Nonlinear Reliability Guardrail}
For nonlinear metrics, the quantitative Poincaré scatter point count ($N_{\text{poincare}}$) directly determines guardrail activation; when $N_{\text{poincare}} < 100$, the system instructs the LLM to rely on time-domain features only, as entropy and DFA estimates become unstable with insufficient data length.

These guardrails use quantitative image-derived features to ensure reliability independent of visual rendering, preventing hallucinations from image scaling or compression artifacts. The guardrails are configurable for ablation studies.

\subsection{Integration and Output (Step 8)}

Step 8 synthesizes all prior step outputs, RAG-retrieved knowledge, and quality warnings into a final structured report. The integration follows a \textbf{hierarchical evidence fusion} strategy:
\begin{enumerate}
    \item \textbf{Collect}: Gather sub-reports from Steps 1--7 plus RAG-retrieved passages with adjusted scores.
    \item \textbf{Conflict Detection}: Identify contradictions (e.g., $\text{sign}(z_{\text{trad}}) \neq \text{sign}(z_{\Delta})$; Step 2 vs.\ Step 6 disagreement).
    \item \textbf{Prompt Assembly}: Inject guardrail warnings, conflict flags, and decision hierarchy reminders into the system prompt.
    \item \textbf{LLM Inference}: Generate structured output under template constraints.
\end{enumerate}

The LLM receives a template-constrained prompt requiring:
\begin{itemize}
    \item Psychophysiological state classification (HVHA/HVLA/LVHA/LVLA)
    \item Confidence level (High/Medium/Low)
    \item Key rationale with evidence citations
    \item Explicit notes on input limitations
\end{itemize}

\textbf{Post-processing.} The system validates output format compliance, detects numerical hallucinations (e.g., MeanHR $>$ 200 bpm), and flags Z-score vs.\ Delta contradictions. The quantitative image features enable cross-validation: if the LLM reports a Poincaré scatter count that differs from the quantitative $N_{\text{poincare}}$, or if PSD band power values are inconsistent with the quantitative features, the system flags potential visual interpretation errors. The system preserves the LLM's original judgment while logging inconsistencies for quality assurance.

\section{Experiments}

\subsection{Dataset and Experimental Setup}

\textbf{Dataset.} We use the DREAMER dataset~\cite{dreamer2018}, comprising 23 participants viewing 18 emotion-eliciting film clips. Each trial provides ECG recordings (256 Hz) for HRV extraction, plus self-reported valence and arousal on a 5-point Likert scale.

\textbf{Label Construction.} Ratings are discretized via median split into four psychophysiological states: HVHA, HVLA, LVHA, and LVLA. Trials with neutral ratings (valence or arousal $= 3$) are excluded from evaluation.

\textbf{Hardware.} All experiments run on NVIDIA RTX 5090 (32GB VRAM).

\subsection{Models and Inference Configuration}

Table~\ref{tab:models} lists the evaluated LLMs. All use the full pipeline (RAG, guardrails, Delta Z-score enabled; EEG disabled). Ablations are conducted on Qwen 8B for its balanced capability and efficiency.

\begin{table}[htbp]
\caption{Model and Inference Configurations}
\begin{center}
\resizebox{\columnwidth}{!}{
\begin{tabular}{|>{\centering\arraybackslash}m{2.6cm}|>{\centering\arraybackslash}m{1.0cm}|>{\centering\arraybackslash}m{1.4cm}|>{\centering\arraybackslash}m{2.8cm}|}
\hline
\textbf{Model} & \textbf{Params} & \textbf{Precision} & \textbf{Notes} \\
\hline
MedGemma 4B & 4B & bfloat16 & Full precision \\
\hline
MedGemma 27B & 27B & 4-bit & Memory-constrained \\
\hline
MedGemma 27B+CoT~\cite{medgemma3thinking} & 27B & 4-bit & Chain-of-thought \\
\hline
Qwen 4B & 4B & bfloat16 & Full precision \\
\hline
Qwen 8B & 8B & bfloat16 & \textbf{Ablation baseline} \\
\hline
\multicolumn{4}{|c|}{\textit{Fixed Inference Parameters}} \\
\hline
\multicolumn{2}{|c|}{Temperature: 0.3} & \multicolumn{2}{c|}{Top-$p$: 0.85} \\
\hline
\multicolumn{2}{|c|}{Repetition penalty: 1.05} & \multicolumn{2}{c|}{Max tokens (Step 8): 4096} \\
\hline
\end{tabular}
}
\label{tab:models}
\end{center}
\end{table}

\subsubsection{Generation Stability Measures}
During development, we identified and mitigated three failure modes:
\begin{itemize}
    \item \textbf{Token repetition collapse}: Addressed via \texttt{repetition\_penalty=1.05} and runtime truncation upon detecting repeated patterns ($>$50 identical characters or $>$3 n-gram repeats).
    \item \textbf{Unicode substitution}: Counterintuitively, \texttt{no\_repeat\_ngram\_size=3} \textit{triggered} Greek letter substitutions (e.g., ``$\alpha$'' for ``a''); disabling it resolved the issue. Examples: incorrect ``par$\alpha$sympathetic'' vs.\ correct ``parasympathetic''; incorrect ``pr$\rho$cessing'' vs.\ correct ``processing''.
    \item \textbf{Numerical hallucinations}: Post-processing validates ranges (MeanHR $\in$ [40, 200] bpm, RMSSD $\in$ [0, 500] ms) and flags implausible values.
\end{itemize}

\subsection{Evaluation Metrics}

Table~\ref{tab:metrics} summarizes our three-tier evaluation framework.

\begin{table}[htbp]
\caption{Evaluation Metrics}
\begin{center}
\resizebox{\columnwidth}{!}{
\begin{tabular}{|>{\centering\arraybackslash}m{0.8cm}|>{\centering\arraybackslash}m{2.6cm}|>{\centering\arraybackslash}m{4.2cm}|}
\hline
\textbf{ID} & \textbf{Metric} & \textbf{Description} \\
\hline
\multicolumn{3}{|c|}{\textit{Task Performance (GT $\neq$ neutral)}} \\
\hline
T1 & GT Accuracy & 4-class exact match \\
\hline
T2 & Arousal Accuracy & High/Low arousal match \\
\hline
T3 & Vagal Accuracy & High/Low vagal match \\
\hline
\multicolumn{3}{|c|}{\textit{Output Quality}} \\
\hline
Q1 & State Fill Rate & State field extracted \\
\hline
Q2 & Valid Label Rate & Valid label (not Unknown/Other) \\
\hline
Q3 & RAG Header Rate & Contains RAG citations \\
\hline
Q4 & Z-score Presence & Includes z-score values \\
\hline
\multicolumn{3}{|c|}{\textit{Cross-Model Consistency}} \\
\hline
C1 & State Agreement & Match vs.\ baseline model \\
\hline
\multicolumn{3}{|c|}{\textit{Affective Space Distance}} \\
\hline
WAD & Weighted Affective Distance & Euclidean distance in Circumplex Model \\
\hline
\end{tabular}
}
\label{tab:metrics}
\end{center}
\end{table}

\textbf{Task metrics} (T1--T3) measure classification accuracy; T1 requires exact 4-class match, T2/T3 evaluate dimensions independently. \textbf{Quality metrics} (Q1--Q4) assess format compliance and evidence attribution. \textbf{Consistency metric} (C1) compares state outputs across configurations.

\textbf{Weighted Affective Distance (WAD)} quantifies prediction errors in the Circumplex Model of Affect. We map each psychophysiological state to a 2D coordinate: HVHA = (+1, +1), HVLA = (+1, -1), LVHA = (-1, +1), LVLA = (-1, -1), where the first dimension represents Valence (High = +1, Low = -1) and the second represents Arousal (High = +1, Low = -1). The affective distance between ground truth $y_{\text{gt}}$ and prediction $y_{\text{pred}}$ is computed as the Euclidean distance:
\begin{equation}
    d(y_{\text{gt}}, y_{\text{pred}}) = \sqrt{(v_{\text{gt}} - v_{\text{pred}})^2 + (a_{\text{gt}} - a_{\text{pred}})^2},
    \label{eq:wad-distance}
\end{equation}
where $v$ and $a$ are the Valence and Arousal coordinates, respectively. The mean WAD across all samples is:
\begin{equation}
    \text{WAD} = \frac{1}{N} \sum_{i=1}^{N} d(y_{\text{gt}}^{(i)}, y_{\text{pred}}^{(i)}),
    \label{eq:mean-wad}
\end{equation}
where $N$ is the number of valid samples (GT $\neq$ neutral). Lower WAD indicates better performance in the affective space. The normalized WAD (0--1 scale) is computed by dividing mean WAD by the maximum possible distance ($\sqrt{8} \approx 2.83$), corresponding to diagonal quadrant errors (e.g., HVHA $\rightarrow$ LVLA).

\subsection{Ablation Study Design}

We conduct component ablations on Qwen 8B (Table~\ref{tab:ablation-config}) to isolate each module's contribution.

\begin{table}[htbp]
\caption{Ablation Configurations}
\begin{center}
\begin{tabular}{|l|c|c|c|}
\hline
\textbf{Setting} & \textbf{RAG} & \textbf{Guardrails} & \textbf{$\Delta$Z} \\
\hline
Full System & \checkmark & \checkmark & \checkmark \\
\hline
w/o RAG & $\times$ & \checkmark & \checkmark \\
\hline
w/o Guardrails & \checkmark & $\times$ & \checkmark \\
\hline
w/o $\Delta$Z & \checkmark & \checkmark & $\times$ \\
\hline
Minimal & $\times$ & $\times$ & $\times$ \\
\hline
\end{tabular}
\label{tab:ablation-config}
\end{center}
\end{table}

\noindent\textbf{Ablation hypotheses:}
\begin{itemize}
    \item \textbf{w/o RAG}: Tests whether evidence grounding reduces hallucinations.
    \item \textbf{w/o Guardrails}: Exposes system to HRV interpretation pitfalls (RSA contamination, unreliable nonlinear metrics).
    \item \textbf{w/o $\Delta$Z}: Tests whether individualized normalization reduces z-score/delta direction conflicts.
\end{itemize}

\section{Results}

\subsection{Dataset Statistics and Evaluation Setup}

We evaluated C-GRASP on the DREAMER dataset, comprising 414 trials from 23 participants. After excluding neutral trials (valence or arousal rating = 3), 233 trials remained for task performance evaluation (T1--T3). The remaining 181 neutral trials were excluded from accuracy calculations but retained for output quality assessment (Q1--Q4) and cross-model consistency analysis (C1).

\subsection{Multi-Model Performance Comparison}

Table~\ref{tab:results-multi-model} compares model performance across task (T1--T3) and quality (Q1--Q4) metrics. MedGemma-Thinking significantly outperformed the baseline and other variants, achieving the highest 4-class accuracy (37.3\%) and balanced predictions across the affective space. While Qwen-V3-4B-it attained comparable T1 scores (36.1\%), this performance is misleading; Table~\ref{tab:prediction-distribution} reveals severe mode collapse, with 91.8\% of samples predicted as HVHA. In contrast, MedGemma-Thinking maintained a physiologically plausible distribution (Table~\ref{tab:prediction-distribution}), confirming that its superior performance stems from genuine reasoning rather than statistical guessing.

Regarding output quality, all models demonstrated high format compliance ($>$88\% state fill rate). MedGemma variants consistently included RAG citations (100\% Q3) and numerical grounding ($>$94\% Q4), validating the effectiveness of the template-constrained generation. Cross-model consistency analysis (C1) highlights that architectural differences significantly impact reasoning pathways, with MedGemma models showing divergent but clinically grounded interpretations compared to the Qwen baseline.

\begin{figure}[htbp]
\centerline{\includegraphics[width=\linewidth]{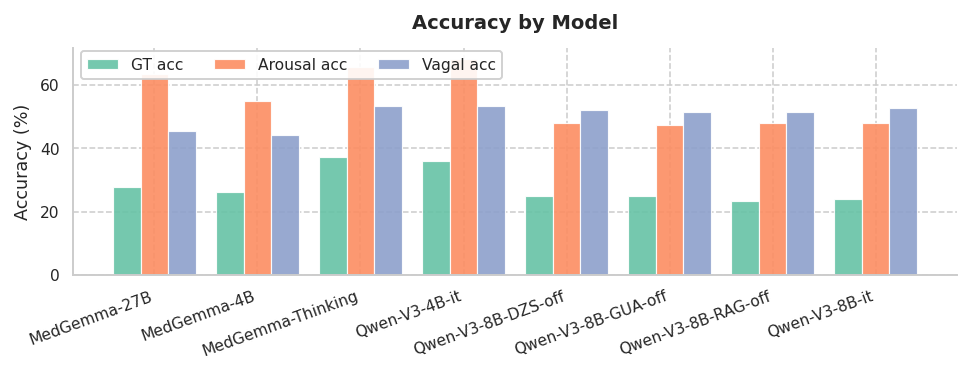}}
\caption{Task performance metrics (T1--T3) across models.}
\label{fig:accuracy}
\end{figure}

Fig.~\ref{fig:accuracy} visualizes these trends, confirming that Arousal classification (T2) is generally more robust than Vagal classification (T3) across architectures, though MedGemma-Thinking maintains the best balance between the two dimensions.

\begin{table}[htbp]
\caption{Multi-Model Performance Comparison}
\begin{center}
\resizebox{\columnwidth}{!}{
\begin{tabular}{|>{\centering\arraybackslash}m{2.0cm}|>{\centering\arraybackslash}m{0.8cm}|>{\centering\arraybackslash}m{0.8cm}|>{\centering\arraybackslash}m{0.8cm}|>{\centering\arraybackslash}m{0.8cm}|>{\centering\arraybackslash}m{0.8cm}|>{\centering\arraybackslash}m{0.8cm}|>{\centering\arraybackslash}m{0.8cm}|>{\centering\arraybackslash}m{0.8cm}|}
\hline
\textbf{Model} & \textbf{T1} & \textbf{T2} & \textbf{T3} & \textbf{Q1} & \textbf{Q2} & \textbf{Q3} & \textbf{Q4} & \textbf{C1} \\
\hline
MedGemma-27B & 27.9 & 63.5 & 45.5 & 99.8 & 99.8 & 100.0 & 99.5 & 19.8 \\
\hline
MedGemma-Thinking & 37.3 & 65.7 & 53.2 & 99.8 & 99.8 & 100.0 & 94.9 & 23.4 \\
\hline
MedGemma-4B & 26.2 & 51.9 & 44.2 & 88.4 & 88.4 & 100.0 & 96.6 & 15.2 \\
\hline
Qwen-V3-4B-it & 36.1 & 68.2 & 53.2 & 100.0 & 100.0 & 0.0$^*$ & 100.0 & 36.5 \\
\hline
Qwen-V3-8B-it & 24.0 & 48.1 & 52.8 & 100.0 & 100.0 & 0.0$^*$ & 100.0 & 100.0 \\
\hline
\end{tabular}
}
\footnotesize
$^*$Qwen models use a different output format without explicit RAG header markers (by design).
\label{tab:results-multi-model}
\end{center}
\end{table}

\begin{table}[htbp]
\caption{Prediction Distribution Analysis: Revealing Mode Collapse}
\begin{center}
\resizebox{\columnwidth}{!}{
\begin{tabular}{|>{\centering\arraybackslash}m{2.5cm}|>{\centering\arraybackslash}m{1.2cm}|>{\centering\arraybackslash}m{1.2cm}|>{\centering\arraybackslash}m{1.2cm}|>{\centering\arraybackslash}m{1.2cm}|>{\centering\arraybackslash}m{1.0cm}|}
\hline
\textbf{Model} & \textbf{HVHA} & \textbf{HVLA} & \textbf{LVHA} & \textbf{LVLA} & \textbf{Total} \\
\hline
MedGemma-Thinking & 189 (45.7\%) & 22 (5.3\%) & 191 (46.1\%) & 11 (2.7\%) & 414 \\
\hline
Qwen-V3-4B-it & \textbf{380 (91.8\%)} & \textbf{0 (0.0\%)} & \textbf{34 (8.2\%)} & \textbf{0 (0.0\%)} & 414 \\
\hline
MedGemma-4B & 40 (9.7\%) & 0 (0.0\%) & 275 (66.4\%) & 51 (12.3\%) & 366$^*$ \\
\hline
Qwen-V3-8B-it & 155 (37.4\%) & 17 (4.1\%) & 6 (1.4\%) & 236 (57.0\%) & 414 \\
\hline
MedGemma-27B & 178 (43.0\%) & 22 (5.3\%) & 204 (49.3\%) & 9 (2.2\%) & 414 \\
\hline
\end{tabular}
}
\footnotesize
$^*$MedGemma-4B has 48 samples with Unknown/Other predictions (not shown in table).
\label{tab:prediction-distribution}
\end{center}
\end{table}

\subsection{Ablation Study Results}

We conducted component ablations on Qwen-V3-8B-it to isolate the contribution of each module (Table~\ref{tab:results-ablation}). While ablation variants showed minor fluctuations in classification accuracy (T1), the removal of any core module---RAG, Guardrails, or $\Delta$Z---resulted in significant deviations in reasoning consistency (C1: 71.5--73.4\% agreement with Full System). This indicates that the complete C-GRASP architecture is essential for stabilizing model outputs, particularly in handling ambiguous physiological signals where single-component systems may drift. Notably, RAG integration appeared to balance multi-dimensional reasoning, as its removal improved Arousal accuracy but degraded Vagal accuracy, suggesting a trade-off between sensitivity and specificity that RAG helps mediate.

\begin{figure}[htbp]
\centerline{\includegraphics[width=\linewidth]{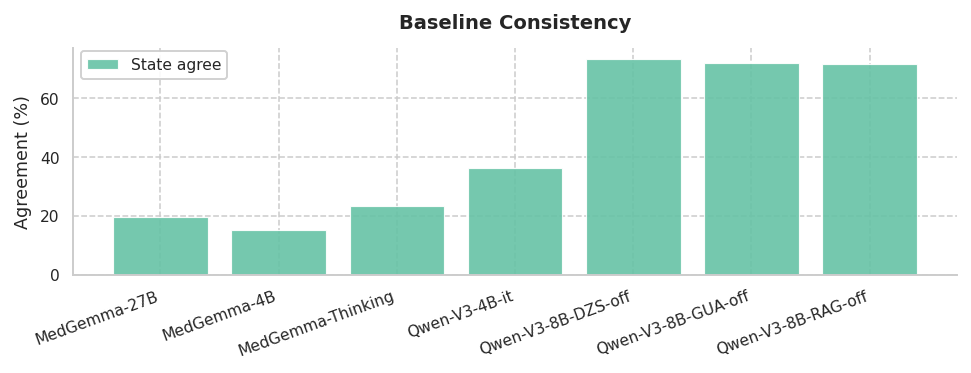}}
\caption{Cross-model consistency (C1: State Agreement) between the full C-GRASP system and ablation variants. The figure compares the percentage of state predictions that match the full system baseline across different component configurations (w/o RAG, w/o Guardrails, w/o $\Delta$Z). Higher values indicate greater output stability and consistency.}
\label{fig:baseline-agreement}
\end{figure}

Fig.~\ref{fig:baseline-agreement} visualizes cross-model consistency (C1, State Agreement) between the full system and ablation variants. The figure shows that removing any component (RAG, Guardrails, or $\Delta$Z) reduces state agreement to approximately 70--73\% compared to the full system (100\%), indicating that all components contribute to output stability. The consistency patterns demonstrate that the complete C-GRASP architecture is essential for maintaining stable predictions across different configurations, particularly when handling ambiguous physiological signals where single-component systems may produce divergent outputs.

\begin{table}[htbp]
\caption{Ablation Study Results (Qwen-V3-8B-it)}
\begin{center}
\resizebox{\columnwidth}{!}{
\begin{tabular}{|>{\centering\arraybackslash}m{2.0cm}|>{\centering\arraybackslash}m{0.8cm}|>{\centering\arraybackslash}m{0.8cm}|>{\centering\arraybackslash}m{0.8cm}|>{\centering\arraybackslash}m{0.8cm}|>{\centering\arraybackslash}m{0.8cm}|>{\centering\arraybackslash}m{0.8cm}|>{\centering\arraybackslash}m{0.8cm}|}
\hline
\textbf{Setting} & \textbf{T1} & \textbf{T2} & \textbf{T3} & \textbf{Q1} & \textbf{Q2} & \textbf{Q4} & \textbf{C1} \\
\hline
Full System & 24.0 & 48.1 & 52.8 & 100.0 & 100.0 & 100.0 & 100.0 \\
\hline
w/o RAG & 23.2 & 48.1 & 49.7 & 100.0 & 100.0 & 100.0 & 71.5 \\
\hline
w/o Guardrails & 24.9 & 47.2 & 51.5 & 100.0 & 100.0 & 100.0 & 72.0 \\
\hline
w/o $\Delta$Z & 24.9 & 48.1 & 51.9 & 100.0 & 100.0 & 100.0 & 73.4 \\
\hline
\end{tabular}
}
\label{tab:results-ablation}
\end{center}
\end{table}

\subsection{Clinical Reasoning Consistency (CRC) Analysis}

We introduced a novel \textbf{Clinical Reasoning Consistency (CRC)} metric to evaluate whether model-generated text descriptions align with the quantitative Z-scores reported in the same output. CRC measures the agreement between Z-score direction (positive/negative) and the presence of corresponding clinical keywords in the generated text.

\textbf{CRC Computation Procedure.} For each model-generated report $R$, we extract Z-scores for seven HRV metrics $\mathcal{M} = \{\text{RMSSD}, \text{SDNN}, \text{pNN50}, \text{MeanHR}, \text{LF/HF}, \text{SampEn}, \text{DFA}_\alpha\}$ using pattern matching. For each metric $m \in \mathcal{M}$ with extracted Z-score $z_m$, we define a consistency check as follows:

Let $\mathcal{K}_m^+$ and $\mathcal{K}_m^-$ denote the sets of positive and negative clinical keywords for metric $m$, respectively. For example, for RMSSD, $\mathcal{K}_{\text{RMSSD}}^+ = \{\text{``increased vagal''}, \text{``elevated parasympathetic''}, \ldots\}$ and $\mathcal{K}_{\text{RMSSD}}^- = \{\text{``reduced vagal''}, \text{``sympathetic dominance''}, \ldots\}$. We count keyword occurrences in the report text $R$:
\begin{equation}
n_m^+ = \sum_{k \in \mathcal{K}_m^+} \mathbb{1}[k \in R], \quad n_m^- = \sum_{k \in \mathcal{K}_m^-} \mathbb{1}[k \in R],
\label{eq:keyword-count}
\end{equation}
where $\mathbb{1}[\cdot]$ is the indicator function.

For each metric $m$ with $|z_m| > \tau$ (where $\tau = 0.5$ is the directional significance threshold), we classify the metric-keyword alignment as:
{\small
\begin{equation}
\text{consistency}_m = \begin{cases}
\text{consistent} & \text{if } (z_m < -\tau \land n_m^- > n_m^+) \lor \\
& \quad (z_m > \tau \land n_m^+ > n_m^-) \\
\text{inconsistent} & \text{if } (z_m < -\tau \land n_m^+ > n_m^-) \lor \\
& \quad (z_m > \tau \land n_m^- > n_m^+) \\
\text{neutral} & \text{if } |z_m| \leq \tau \\
\text{no\_keywords} & \text{if } n_m^+ = 0 \land n_m^- = 0
\end{cases}
\label{eq:consistency-check}
\end{equation}
}

The CRC score for report $R$ is then computed as:
{\small
\begin{equation}
\text{CRC}(R) = \frac{N_{\text{consistent}}}{N_{\text{consistent}} + N_{\text{inconsistent}}},
\label{eq:crc}
\end{equation}
}
where the denominator excludes neutral and no-keyword cases. The overall CRC across all reports is the mean of per-report CRC scores. This formulation ensures that CRC measures the alignment between quantitative Z-scores (from Steps 1--7) and qualitative clinical descriptions (in Step 8), providing a traceable link between the reasoning chain and the final output.

\begin{figure}[htbp]
\centerline{\includegraphics[width=\linewidth]{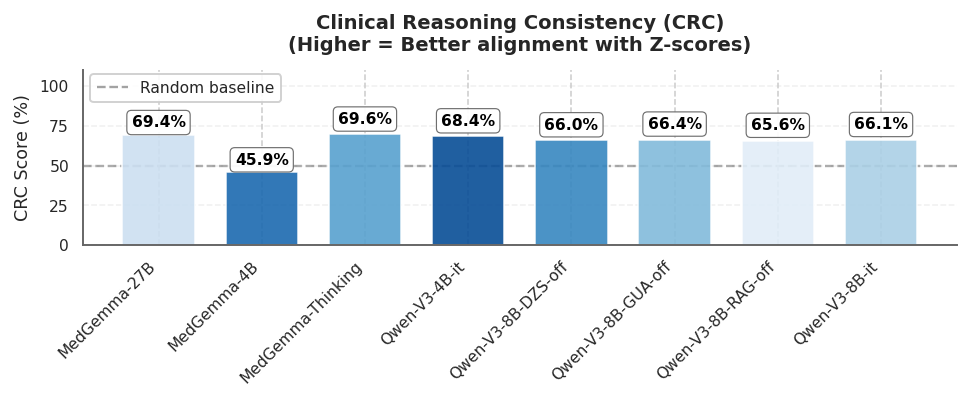}}
\caption{Clinical Reasoning Consistency (CRC) scores across models.}
\label{fig:crc-score}
\end{figure}

Fig.~\ref{fig:crc-score} summarizes CRC across models, showing scores concentrated in the mid-to-high 60s and highlighting MedGemma-Thinking as the strongest performer on numerical-textual alignment.

Ablation analysis showed that removing guardrails (w/o Guardrails) slightly improved CRC to 66.4\%, while removing RAG (w/o RAG) maintained similar CRC (65.6\%) and removing $\Delta$Z (w/o $\Delta$Z) achieved 66.0\%. This suggests that guardrails may occasionally constrain reasoning in ways that reduce numerical-textual consistency, though the differences are small ($<$2 percentage points).

MedGemma-4B showed notably lower CRC (45.9\%) with fewer total checks (196 vs.\ $>$1000 for other models), indicating that smaller models may struggle to maintain consistent numerical-textual alignment, possibly due to limited context capacity or weaker numerical reasoning capabilities.

Overall, most models cluster within a narrow CRC band (roughly 66--70\%), indicating that numerical-textual alignment is generally maintained but still leaves room for improvement in fully resolving residual inconsistencies.

% (Page-limit optimization) Removed CRC distribution figure and detailed CRC table; Fig.~\ref{fig:crc-score} provides the primary summary.

\subsection{F1-Score Analysis for Class Imbalance Robustness}

\begin{figure}[htbp]
\centerline{\includegraphics[width=\linewidth]{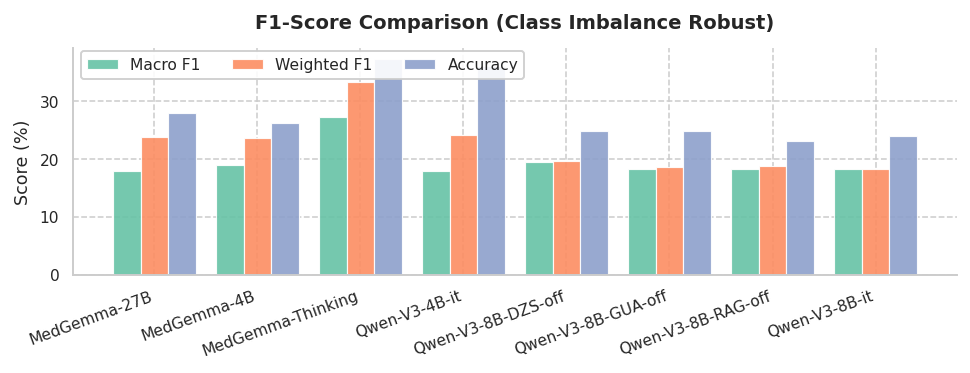}}
\caption{Macro and Weighted F1 across models.}
\label{fig:f1-scores}
\end{figure}

Given the class imbalance in the DREAMER dataset, we report Macro F1 and Weighted F1 scores to provide a more robust evaluation than accuracy alone. Fig.~\ref{fig:f1-scores} compares F1 metrics across models. MedGemma-Thinking achieved the highest Macro F1 (27.3\%) and Weighted F1 (33.4\%), outperforming its accuracy (T1: 37.3\%), indicating that while it achieves high accuracy, it may still struggle with minority classes. The baseline Qwen-V3-8B-it achieved Macro F1 of 18.2\% and Weighted F1 of 18.3\%, both lower than its accuracy (24.0\%), suggesting that its predictions may be biased toward majority classes.

These aggregate F1 metrics are more stable than accuracy under class imbalance and provide a concise view of model robustness.

% (Page-limit optimization) Removed per-class F1 figure; Macro/Weighted F1 in Fig.~\ref{fig:f1-scores} provides the main summary.

\subsection{Error Pattern Analysis}

Analysis of confusion matrices revealed that models showed higher confusion between adjacent quadrants (e.g., HVHA vs.\ HVLA) than between diagonal quadrants (e.g., HVHA vs.\ LVLA), consistent with the Circumplex Model of Affect. Overall, most models show higher Arousal confusion than Valence confusion, and the ablation variants (w/o RAG, w/o Guardrails, w/o $\Delta$Z) follow similar patterns to the full system, suggesting that these components primarily affect output stability rather than fundamental error patterns.

The full system (Qwen-V3-8B-it) showed balanced error distribution across dimensions (110 Valence, 121 Arousal), while ablation variants showed slight shifts: w/o RAG showed increased Valence confusion (113), w/o Guardrails showed increased Arousal confusion (123), and w/o $\Delta$Z showed minimal change (112 Valence, 121 Arousal), suggesting that RAG helps stabilize Valence reasoning while guardrails help stabilize Arousal reasoning.

\begin{figure}[htbp]
\centerline{\includegraphics[width=\linewidth]{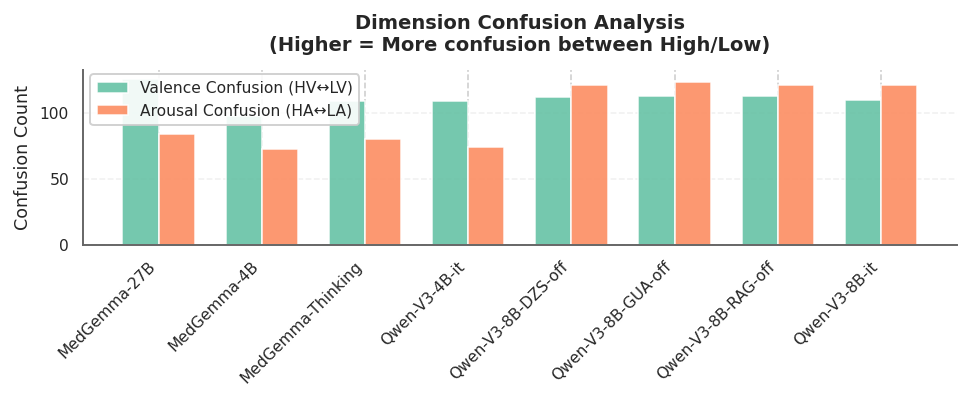}}
\caption{Dimension confusion counts (Valence vs.\ Arousal) across models.}
\label{fig:dimension-confusion}
\end{figure}

Fig.~\ref{fig:dimension-confusion} visualizes dimension-level confusion patterns, confirming that Arousal confusion is generally higher than Valence confusion across most models, with the exception of MedGemma-4B and Qwen-V3-4B-it which show higher Valence confusion.

\begin{figure}[htbp]
\centerline{
\begin{minipage}{0.48\linewidth}
\centering
\adjustbox{width=\textwidth,height=5.0cm,center}{\includegraphics{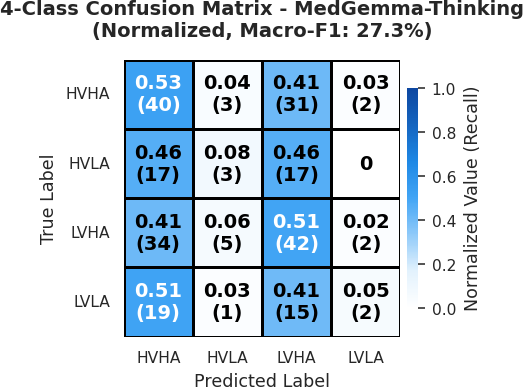}}
\caption*{(a) MedGemma-Thinking}
\end{minipage}
\hfill
\begin{minipage}{0.48\linewidth}
\centering
\adjustbox{width=\textwidth,height=5.0cm,center}{\includegraphics{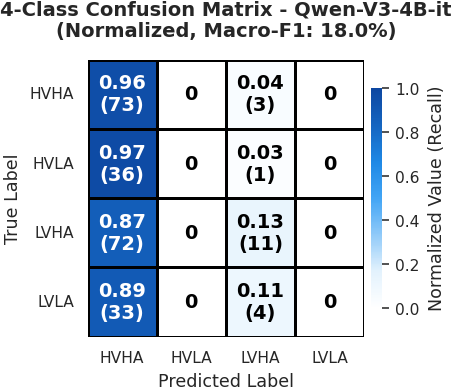}}
\caption*{(b) Qwen-V3-4B-it}
\end{minipage}
}
\caption{Confusion matrices: (a) MedGemma-Thinking, (b) Qwen-V3-4B-it.}
\label{fig:confusion-comparison}
\end{figure}

Fig.~\ref{fig:confusion-comparison} provides a striking visual comparison of confusion matrices, revealing the mode collapse in Qwen-V3-4B-it. The left panel (MedGemma-Thinking) shows a balanced confusion pattern with predictions distributed across all four quadrants, demonstrating the model's ability to capture the full affective spectrum. In contrast, the right panel (Qwen-V3-4B-it) exhibits severe mode collapse: the model predicts HVHA for nearly all samples (380/414), resulting in a confusion matrix dominated by the HVHA column. This visualization clearly demonstrates that high accuracy metrics can mask fundamental model limitations when prediction distributions are not examined.

Across models, most errors occur between adjacent quadrants (sharing either Valence or Arousal), aligning with the Circumplex Model where neighboring affective states are inherently more confusable than opposite states.

Table~\ref{tab:wad-results} reports the Weighted Affective Distance (WAD) analysis across all models. MedGemma-Thinking achieved the lowest mean WAD (1.41), followed by Qwen-V3-4B-it (1.40) and MedGemma-4B (1.52). The normalized WAD (0--1 scale) shows MedGemma-Thinking and Qwen-V3-4B-it both achieving approximately 0.50, indicating moderate performance in the affective space. The ablation variants (w/o RAG, w/o Guardrails, w/o $\Delta$Z) showed higher WAD values (1.71--1.73), suggesting that all components contribute to reducing affective distance errors.

Error decomposition reveals that most models show a higher proportion of correct predictions than errors. MedGemma-Thinking achieved 87 correct predictions (37.3\%) with balanced error distribution (Valence errors: 66, Arousal errors: 37, Cross-quadrant errors: 43). In contrast, the ablation variants showed fewer correct predictions (54--58) and higher cross-quadrant error rates, indicating that the full system better captures the affective spectrum.

\begin{figure}[htbp]
\centerline{\includegraphics[width=\linewidth]{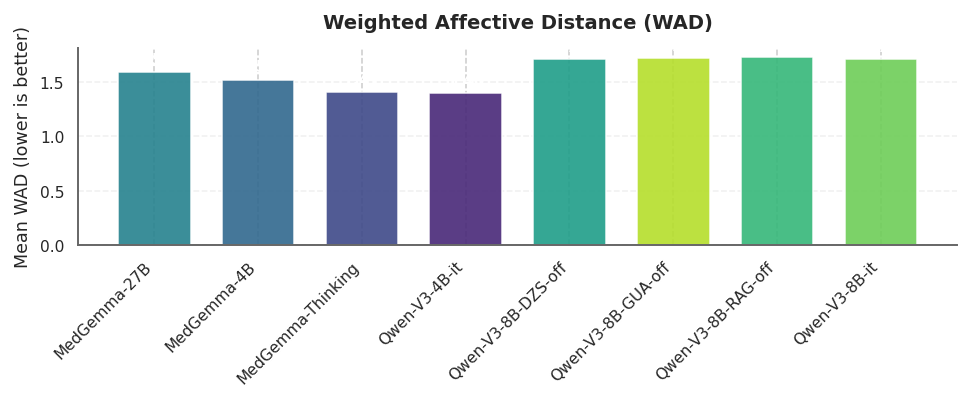}}
\caption{Weighted Affective Distance (WAD) across models.}
\label{fig:wad-score}
\end{figure}

Fig.~\ref{fig:wad-score} visualizes mean WAD and normalized WAD across models, confirming that MedGemma-Thinking and Qwen-V3-4B-it achieve the best performance in the affective space. Across models, cross-quadrant errors (both dimensions wrong) remain the least frequent, consistent with the confusion matrix analysis.

% (Page-limit optimization) Removed dimension-level confusion summary table; key trend is described in text and visualized in Fig.~\ref{fig:dimension-confusion}.

% (Page-limit optimization) Removed baseline confusion-matrix figure; adjacent-quadrant confusion pattern is summarized in text.

\begin{table}[htbp]
\caption{Weighted Affective Distance (WAD) Analysis}
\begin{center}
\resizebox{\columnwidth}{!}{
\begin{tabular}{|>{\centering\arraybackslash}m{2.0cm}|>{\centering\arraybackslash}m{1.0cm}|>{\centering\arraybackslash}m{1.0cm}|>{\centering\arraybackslash}m{0.8cm}|>{\centering\arraybackslash}m{0.8cm}|>{\centering\arraybackslash}m{0.8cm}|>{\centering\arraybackslash}m{0.8cm}|>{\centering\arraybackslash}m{0.8cm}|}
\hline
\textbf{Model} & \textbf{Mean WAD} & \textbf{Norm. WAD} & \textbf{Correct} & \textbf{Val. Err} & \textbf{Aro. Err} & \textbf{Both Err} & \textbf{Total} \\
\hline
MedGemma-Thinking & 1.41 & 0.50 & 87 & 66 & 37 & 43 & 233 \\
\hline
Qwen-V3-4B-it & 1.40 & 0.49 & 84 & 75 & 40 & 34 & 233 \\
\hline
MedGemma-4B & 1.52 & 0.54 & 61 & 67 & 42 & 31 & 201 \\
\hline
MedGemma-27B & 1.59 & 0.56 & 65 & 83 & 41 & 43 & 232 \\
\hline
Qwen-V3-8B-it & 1.71 & 0.61 & 56 & 56 & 67 & 54 & 233 \\
\hline
w/o RAG & 1.73 & 0.61 & 54 & 58 & 66 & 55 & 233 \\
\hline
w/o Guardrails & 1.72 & 0.61 & 58 & 52 & 62 & 61 & 233 \\
\hline
w/o $\Delta$Z & 1.71 & 0.60 & 58 & 54 & 63 & 58 & 233 \\
\hline
\end{tabular}
}
\footnotesize
Val. Err = Valence-only errors; Aro. Err = Arousal-only errors; Both Err = Cross-quadrant errors.
\label{tab:wad-results}
\end{center}
\end{table}

% (Page-limit optimization) Removed WAD error-type distribution figure; key takeaway is summarized in text.

\section{Discussion and Conclusion}
C-GRASP targets clinically traceable affective signal interpretation by decomposing HRV reasoning into auditable steps with quantitative guardrails and individualized normalization, aiming to reduce known failure modes such as RSA-confounded frequency indices~\cite{billman2013}, short-window nonlinear instability~\cite{zhao2015}, and population-norm bias. In our DREAMER evaluation~\cite{dreamer2018}, this design emphasizes conservative metric validity (e.g., RSA-aware gating and data-length gating) and prioritizes within-subject change (Delta Z-score) over absolute thresholds, which is aligned with wearable-style longitudinal monitoring rather than one-off classification.

\subsection{Data Leakage Considerations and Wearable-Style Longitudinal Simulation}
A key limitation is baseline construction in offline evaluation: subject-specific baselines are estimated using all available trials to approximate longitudinal history, which is non-causal because the current (and potentially future) trials can influence the baseline distribution. Although no label information is used, this may slightly overestimate the strength of within-subject evidence, so reported performance should be interpreted as a retrospective upper bound for baseline availability. In real-world deployment, baselines should be computed causally using only prior measurements (e.g., chronological accumulation or leave-one-trial-out baselines) to fully remove this concern.

\subsection{Why Guardrailed RAG + LLM Reasoning Instead of End-to-End Deep Learning}
Compared with end-to-end deep learning, the guardrailed RAG stepwise design is chosen for reliability under heterogeneous wearable conditions (artifact, dropout, protocol differences) and for debuggability: Step 1 quantifies signal quality, Step 3 explicitly restricts frequency-domain interpretation under RSA, and Step 4 restricts nonlinear metrics under insufficient data length. Ablations further indicate that removing RAG, guardrails, or individualized Z-score modules degrades reasoning stability, supporting the view that clinical usefulness depends on traceability and conservative handling of invalid metrics rather than accuracy alone.

\subsection{Quantitative Overview of Model Reasoning}

Table~\ref{tab:model_comparison} presents a comparison of reasoning performance between the representative 4B model (MedGemma-4B-it) and the 27B model (MedGemma3-thinking) under the C-GRASP framework. While Qwen-V3-4B-it is also evaluated in this study, we select MedGemma-4B-it for this detailed qualitative comparison to highlight the specific architectural differences within the same model family (Gemma 2 base) across scales.

\begin{table}[htbp]
\caption{Reasoning Performance Comparison: 4B vs. 27B Models under C-GRASP}
\label{tab:model_comparison}
\centering
\resizebox{\columnwidth}{!}{
\begin{tabular}{lp{3.5cm}p{4.2cm}}
\toprule
\textbf{Evaluation Dimension} & \textbf{MedGemma-4B (SLM)} & \textbf{MedGemma-Thinking (LLM)} \\
\midrule
\textbf{Instruction Following} & Good, but prone to truncation in complex decisions & Excellent, fully executes 8-step reasoning chain \\
\textbf{Conflict Handling} & Tends to prioritize single metrics (e.g., $z_{\text{MeanHR}}$) & Capable of cross-metric Evidence Weighting \\
\textbf{Stability} & Occasional Repetition Collapse & Highly stable, semantically coherent \\
\textbf{Clinical Attribution} & Primarily descriptive labeling & Capable of literature association and physiological mechanism analysis \\
\textbf{CRC Metric*} & 45.9\% & 69.6\% \\
\bottomrule
\multicolumn{3}{l}{\textit{*CRC: Clinical Reasoning Consistency}}
\end{tabular}
}
\end{table}

\subsection{Case-by-Case Analysis}

\subsubsection{Case S03-T01: Nonlinear Metric Conflict and RSA Handling}
\textbf{Data Features:} $z_{\text{SDNN}} = +2.64$ (significantly increased), $z_{\text{SampEn}} = -2.47$ (significantly decreased), accompanied by severe RSA interference.

\textbf{4B Model Performance:} Although MedGemma-4B successfully detected RSA and intercepted LF/HF energy, it exhibited \textbf{``Label Repetition Collapse''}. When weighing SDNN against SampEn, the model became trapped by the low complexity metric, judging the state as ``Dysregulation/Anxiety (LVHA)''.

\textbf{27B CoT Model Performance:} MedGemma3-thinking demonstrated superior decision logic. In its chain-of-thought, it explicitly noted that while SampEn decreased, the data length ($N=231$) was marginal, and priority should be given to the adaptive capacity represented by high variability (SDNN). The model successfully judged the state as ``Focus/Flow (HVHA)'', matching the original label and demonstrating an understanding of metric reliability weights.

\subsubsection{Case S05-T01: Directional Mismatch between Z-Score and Delta}
\textbf{Data Features:} $z_{\text{RMSSD}} = -0.83$ (lower than population norm), but $\Delta \text{RMSSD} = +0.20$ (increased relative to individual baseline).

\textbf{4B Model Performance:} Exhibited \textbf{``Statistical Hypersensitivity''}. The model over-interpreted $z_{\text{MeanHR}} = +1.04$ as indicating high arousal, ignoring the calm state implied by the subject's extremely low respiratory rate (9.4 bpm), leading to a misclassification as LVHA.

\textbf{27B CoT Model Performance:} The model successfully triggered the \texttt{Z-SCORE\_DELTA\_MISMATCH} guardrail. MedGemma3-thinking actively ``corrected'' the misleading nature of the Z-score during reasoning, stating: ``Although the absolute Z-score for RMSSD is negative, the Delta direction indicates improvement relative to the subject's baseline.'' This \textbf{``Individual Trend Priority''} reasoning path is a core academic contribution of the C-GRASP framework, proving that large models can understand the clinical significance of dynamic baselines.

\subsubsection{Case S16-T01: Evidence Weighting at Data Boundaries}
\textbf{Data Features:} $z_{\text{SDNN}} = -0.58$ (low absolute value), $\Delta \text{SDNN} = +4.57$ (significant upward trend).

\textbf{4B Model Performance:} Reasoning was fragmented. While it annotated RSA, it lacked clear logical support in the final emotion mapping.

\textbf{27B CoT Model Performance:} Demonstrated strong \textbf{``Evidence Governance''}. The model explicitly discussed how the subject exhibited autonomic nervous system resilience through a positive current trend (Positive Delta), despite overall HRV levels being below the population average. Although the final label deviated slightly from the subject's subjective feeling (HA vs LA), the reasoning process was based on a rational resolution of the Z-score vs. Delta conflict rather than random guessing.

\subsection{Discovery of Major Logical Failure: The Label Mapping Paradox}
Upon examining additional diagnostic reports, we observed a safety-critical ``label mapping paradox,'' where free-text reasoning indicates high-arousal/stress while the final structured state maps to low arousal, implying a mapping/execution defect rather than purely physiological misunderstanding. This motivates a lightweight Step 8 mapping guardrail: enforce an explicit arousal/valence decision line, apply a deterministic validator to detect text--label contradictions, and trigger constrained regeneration or downgrade confidence when inconsistencies are detected.

\subsection{Future Work: Towards an Autonomous Clinical Reasoning Agent}
To transcend the limitations of unimodal signal classification, future work will focus on constructing a \textbf{Universal Multimodal Clinical Reasoning Agent}. This framework serves not merely as a tool for HRV analysis but as a cornerstone for future autonomous medical monitoring systems:

1) \textbf{Orchestrating Mature Diagnostic Tools:} We aim to extend the system to integrate diverse clinical modalities. The LLM will function as an orchestration core, dynamically coordinating existing medical image classifiers (such as our previous ViT-Hybrid model for EEG~\cite{cheng2025}) with the C-GRASP individualized numerical monitoring module. By synthesizing visual attention heatmaps with numerical baselines, the agent can identify subtle ``image-physiological inconsistencies'' often missed by traditional algorithms.

2) \textbf{Longitudinal Personal Health Profiling:} Leveraging the \textbf{Dual Z-score Priority Hierarchy}, the system will evolve from cross-sectional sampling to long-term \textbf{Digital Twin} monitoring. This allows for dynamic adjustment based on a patient's historical health trajectory, facilitating precise preventive medicine rather than retrospective diagnosis.

3) \textbf{Expanding Clinical Reasoning Consistency:} We will deepen the Clinical Reasoning Consistency (CRC) evaluation framework to establish an automated safety layer. A key objective is to develop enhanced validation mechanisms that detect the ``Label Mapping Paradox,'' ensuring that every AI-generated recommendation is traceable to specific data anomalies, thereby providing clinicians with transparent, evidence-based decision support.

\section*{Acknowledgment}
We thank National Central University (NCU) BorgLab for hardware support. We acknowledge the use of Cursor IDE for code integration, Perplexity AI for literature search in Related Work and RAG corpus construction, and Gemini for translation assistance, writing suggestions, and grammar correction.

\end{document}